\documentclass[10pt,conference]{IEEEtran}

\usepackage{cite}

\ifCLASSINFOpdf
   \usepackage[pdftex]{graphicx}
   \graphicspath{{figs/}}
   \DeclareGraphicsExtensions{.pdf,.jpeg,.png}
\else
   \usepackage[dvips]{graphicx}
   \graphicspath{{../figs/}}
   \DeclareGraphicsExtensions{.eps}
\fi
\usepackage[cmex10]{amsmath}
\usepackage{amsthm}

\usepackage{algorithmic}

\usepackage{array}

\ifCLASSOPTIONcompsoc
  \usepackage[caption=false,font=normalsize,labelfont=sf,textfont=sf]{subfig}
\else
  \usepackage[caption=false,font=footnotesize]{subfig}
\fi
\usepackage{url}

\usepackage{siunitx}
\usepackage[utf8]{inputenc}

\usepackage{chngpage}
\usepackage{algorithm}
\usepackage{multirow}
\usepackage{booktabs}
\usepackage{amsmath,amssymb}

\usepackage[acronym]{glossaries}
\usepackage{balance}
\usepackage[normalem]{ulem} 
\usepackage[colorlinks=true,pagebackref=false,urlcolor=black,linkcolor=black,citecolor=black]{hyperref} 
\usepackage{csquotes}
\usepackage{color, colortbl}
\definecolor{LightCyan}{rgb}{0.88,1,1}
\definecolor{lightblue}{rgb}{0.6, 0.81, 0.93}

\usepackage[usenames,dvipsnames]{xcolor}

\begin{document}
%
\title{The Impact of Preprocessing on Deep Representations for Iris Recognition on Unconstrained Environments}

\newif\iffinal
\finaltrue
\newcommand{\jemsid}{74}


\iffinal



%
\author{
	\IEEEauthorblockN{
    	Luiz A. Zanlorensi\IEEEauthorrefmark{1},
		Eduardo Luz\IEEEauthorrefmark{2},
        Rayson Laroca\IEEEauthorrefmark{1},
   	    Alceu S. Britto Jr.\IEEEauthorrefmark{3},
		Luiz S. Oliveira\IEEEauthorrefmark{1},
        David Menotti\IEEEauthorrefmark{1}} 

	\IEEEauthorblockA{
    	\IEEEauthorblockA{\IEEEauthorrefmark{1}Department of Informatics, Federal University of Paran\'a (UFPR), Curitiba, PR, Brazil}
    	\IEEEauthorrefmark{2}Computing Department, Federal University of Ouro Preto (UFOP), Ouro Preto, MG, Brazil}
    	\IEEEauthorblockA{\IEEEauthorrefmark{3}Postgraduate Program in Informatics, Pontifical Catholic University of Paran\'a (PUCPR), Curitiba, PR, Brazil}
}

\else
  \author{SIBGRAPI paper ID: \jemsid \\ }
\fi

\maketitle

\newacronym{cnn}{CNN}{Convolutional Neural Network}
\newacronym{eer}{EER}{Equal Error Rate}
\newacronym{far}{FAR}{False Acceptance Rate}
\newacronym{frr}{FRR}{False Rejection Rate}
\newacronym{nir}{NIR}{near-infrared}
\newacronym{pfb}{PFB}{pairwise filter bank}
\newacronym{svm}{SVM}{Support Vector Machine}

\begin{abstract}
The use of iris as a biometric trait is widely used because of its high level of distinction and uniqueness.
Nowadays, one of the major research challenges relies on the recognition of  iris images obtained in visible spectrum under unconstrained  environments.
In this scenario, the acquired iris are affected by  capture distance, rotation, blur, motion blur, low contrast and specular reflection, creating noises that disturb the iris recognition systems.
Besides delineating the iris region, usually preprocessing techniques such as normalization and segmentation of noisy iris images are employed to minimize these problems. 
But these techniques inevitably run into some errors.
In this context, we propose the use of deep representations, more specifically,  architectures based on VGG and ResNet-50 networks, for dealing  with the images using (and not)  iris segmentation and normalization.
We use transfer learning from the face domain and also propose a specific data augmentation technique for iris images.
Our results show that the approach using non-normalized and only circle-delimited iris images reaches a new state of the art in the official protocol of the NICE.II competition, a subset of the UBIRIS database, one of the most challenging databases on unconstrained environments, reporting an average \gls*{eer} of 13.98\% which represents an absolute reduction of about~5\%.
\end{abstract}


\IEEEpeerreviewmaketitle

\section{Introduction}
\label{sec:intro}

\glsresetall





Biometrics has many applications such as verification, identification, duplicity verification, which makes it an important research area.
A biometric system basically consists of extracting and matching distinctive features from a person.
These patterns are stored as a new sample which is subsequently used in the process of comparing and determining the identity of each person within a population.
Considering that biometric systems require robustness combined with high accuracy, the methods applied to identify individuals are in constant development. 

Biometric methods that identify people based on their physical or behavioral features are interesting due to the fact that a person cannot lose or forget its physical characteristics, as can occur with other means of identification such as passwords or identity cards~\cite{Bowyer2008}.
The use of eye traits becomes interesting because it provides a framework for non-invasive screening technology.
Another important factor is that biomedical literature suggests that irises are as distinct as other biometric sources such as fingerprints or patterns of retinal blood vessels~\cite{Wildes1997}.

Research using iris images obtained in \gls*{nir} showed very promising results and reported low rates of recognition error~\cite{Bowyer2008, Phillips2008}.
Currently, one of the greatest challenges in iris recognition is the use of images obtained in visible spectrum under uncontrolled environments~\cite{Proenca2012, DeMarsico2017}.
The main difficulty in iris recognition using these images is that they may have problems such as noise, rotation, blur, motion blur, low contrast, specular reflection, among others.
Generally, techniques such as normalization~\cite{Daugman1993} and segmentation~\cite{Proenca2012} are applied to correct or reduce these problems.

\begin{figure}[!tb]
\begin{center}
 \includegraphics[width=.95\linewidth]{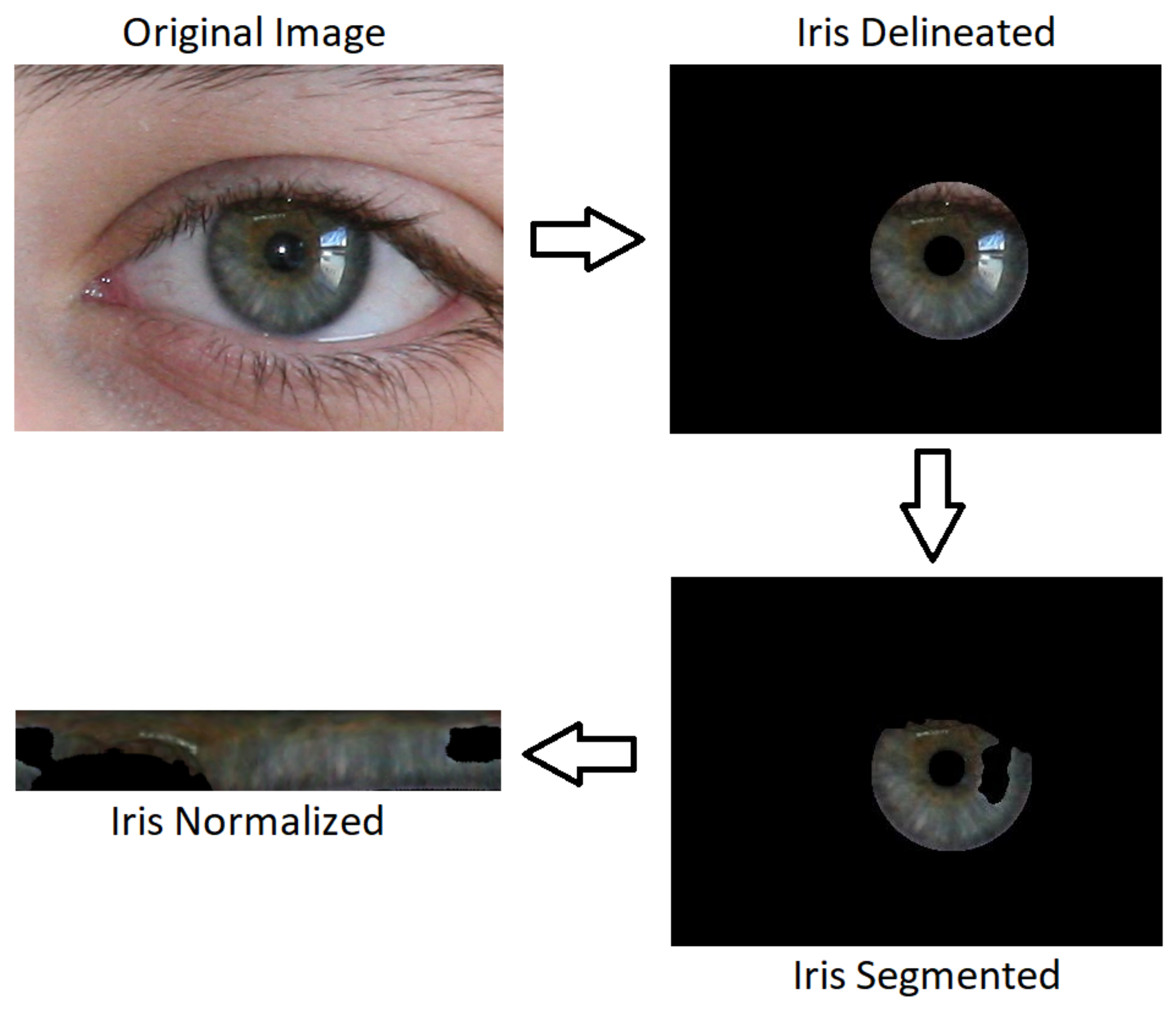}
\end{center}  
\vskip-6pt
\caption{Iris preprocessing stages. }
\label{fig:preprocess}
\end{figure}

With the recent development of deep learning, some applications of this methodology in periocular~\cite{Luz2017, Proenca2018} and iris recognition~\cite{Liu2016, Gangwar2016, Al-Waisy2017, Nguyen2018, Proenca2017} have been performed with interesting results being reported.
The main problem with deep network architectures when trained in small databases is the lack of enough data for generalization usually producing over-fitted models.
One solution to this problem is the use of transfer learning~\cite{Yosinski2014}. 
Another action is the use of data augmentation techniques~\cite{Simard2003, Ciregan2012}. 

In this context, the goal of this work is to evaluate the impact of image preprocessing for deep iris representations acquired on uncontrolled environments.

We evaluated the following preprocesses: iris delineation, iris segmentation for noise removal and iris normalization. The delineation process defines the outer and inner boundaries of the iris (i.e., the pupil). The segmentation process removes the regions where the iris is occluded. Finally, the normalization process consists of transforming the circular region of the iris from the Cartesian space into the Polar one resulting in a rectangular region. 


To improve generalization and avoid overfitting, two \gls*{cnn} models trained for face recognition were fine-tuned and then used as iris representation (or features).
From the comparison of the obtained results,  we can observe that the approaches using only delineated but non-normalized and non-segmented iris image as input for the networks generated new state-of-the-art results for the official protocol of the NICE.II competition.
We also  evaluate the impact of the non-delineated iris images, the squared bounding box.
We chose the NICE.II competition, which is composed as a subset of UBIRIS.v2 database, to evaluate our proposal because it is currently one of the most challenging databases for iris recognition on uncontrolled environments, presenting problems such as noise and different image resolutions.

Since iris recognition may refer to both identification and verification problems, it is important to point out that this paper addresses the verification problem, i.e., to verify if two images are from the same subject. 
Moreover, the experiments are performed on the NICE.II competition protocol which is a verification problem. 


The remainder of this work is organized as follows.
In Section~\ref{sec:related}, we described the methodologies that achieved the best results in NICE.II so far and works that employed deep learning for iris recognition.
Section~\ref{sec:protocol} describes the protocol and database of NICE.II, and the metrics (i.e., \gls*{eer} and decidability) used in our experiments for comparison.
Section~\ref{sec:approach} presents and describes how the experiments were performed.
The results are presented and discussed in Section~\ref{sec:results}, while the conclusions are given in Section~\ref{sec:conclusion}.

\section{Related Work}
\label{sec:related}


In this section, we briefly survey the best performing works in the NICE.II competition and then describe deep learning approaches for iris recognition.

\subsection{The NICE.II competition}

Among eight participants ranked in the NICE.II competition, six of them proposed methodologies considering only iris modality~\cite{Wang2012, Shin2012, Li2012a, DeMarsico2012, Li2012b, Szewczyk2012}. The other two fused iris and periocular region modalities~\cite{Tan2012, Santos2012}. All these  approaches were summarized in the paper \enquote{The results of the NICE.II Iris biometrics competition}~\cite{Bowyer2012} which appeared in 2012. 

Taking advantage of periocular and iris information, the winner~\cite{Tan2012} of the NICE.II reported a decidability value of~$2.57$.
Their methodology consists of image preprocessing, feature extraction from the iris and periocular region, matching of iris and periocular region data, and fusion of the matching results.
For iris feature extraction, ordinal measures and color histogram were used, while texton histogram and semantic information were employed for the periocular region.
The matching scores were obtained using SOBoost learning~\cite{HE2008} for ordinal measures, diffusion distance for color histogram, chi-square distance for texton histogram, and exclusive-or logical operator for semantic label.
The approaches were combined through the sum rule at the matching level.

Using only features extracted from the iris, the best result in NICE.II achieved a decidability value of~$1.82$ and $EER$ of~$19\%$, ranking second in the competition~\cite{Wang2012}.
In that work, Wang et al. first performed the segmentation and normalization of the iris images using the methodology proposed by Daugman~\cite{Daugman1993}.
Then, according to the normalization, irises were partitioned into different numbers of patches.
With the partitions, features were extracted using Gabor filters. Finally, the AdaBoost algorithm was applied to select the best features and compute the similarity.
To the best of our knowledge, this is the state of the art on the NICE.II database when only the iris is used as the biometric trait.


Another work using only iris features for recognition with images obtained in visible wavelength and in unconstrained environments is presented in~\cite{Andersen2017}.
Their methodology consists of four steps: segmentation, normalization, feature extraction and matching with weighted score-level fusion.
Features such as wavelet transform, keypoint-based, generic texture descriptor and color information were extracted and combined.
Results were presented in two subsets of UBIRIS.v2 and MobBIO databases, reporting an \gls*{eer} of $22.04\%$ and $10.32\%$, respectively.
Even though this result is worse than the best one obtained in the NICE.II, which is a subset of UBIRIS.v2 database, a direct comparison is not fair because the data used might be different.

\subsection{Deep learning in iris recognition}
Deep learning is one of the most recent and promising machine learning techniques~\cite{DeMarsico2016}.
Thus, it is natural that there are still a few works that use and apply this technique in iris images.
We describe some of these works below.

The first work applying deep learning to iris recognition was the DeepIris framework,  proposed by Liu et al.~\cite{Liu2016} in 2016.
Their method was applied to the recognition of heterogeneous irises, where the images were obtained with different types of sensors, i.e., the cross-sensor scenario.
The major challenge in this field is the existence of a great intra-class variation, caused by sensor-specific noises.
Thus, handcrafted features generally do not perform well in this type of database.
The study proposed a framework based on deep learning for verification of heterogeneous irises, which establishes the similarity between a pair of iris images using \glspl*{cnn} by learning a bank of pairwise filters.
This methodology differs from the handcrafted feature extraction, since it allows direct learning of a non-linear mapping function between pairs of iris images and their identity with a \gls*{pfb} from different sources.
Thereby, the learned \glspl*{pfb} can be used for new and also different data/subjects, since the learned function is used to establish the similarity between a pair of images.
The experiments were performed in two databases: Q-FIRE and CASIA cross sensor.
Promising results were shown to be better than the baseline methodology, with \gls*{eer} of $0.15$\% in Q-FIRE database and \gls*{eer} of $0.31$\% in CASIA cross-sensor database.


Gangwar \& Joshi~\cite{Gangwar2016} also developed a deep learning application for iris recognition in images obtained from different sensors, called DeepIrisNet.
In their study, a \gls*{cnn} was used to extract features and representations of iris images.
Two databases were used in the experiments: ND-iris-0405 and ND-CrossSensor-Iris-2013. In addition, two \gls*{cnn} architectures were presented, namely DeepIrisNet-A and DeepIrisNet-B.
The former is based on standard convolutional layers, containing $8$~convolutional layers, $8$~normalization layers, and $2$~dropout layers.
DeepIrisNet-B uses inception layers~\cite{Szegedy2015} and its structure consists of $5$ layers of convolution, $7$ of normalization, $2$ of inception and $2$ of dropout.
The results presented five comparisons: effect of segmentation, image rotation analysis, input size analysis, training size analysis and network size analysis.
The proposed methodology demonstrated better robustness compared to the baseline.


The approach proposed by Al-Waisy et al.~\cite{Al-Waisy2017} consists of a multi-biometric iris identification system, using both left and right irises from a person.
Experiments were performed in databases of \gls*{nir} images obtained in controlled environments.
The process has five steps: iris detection, iris normalization, feature extraction, matching with deep learning and, lastly, the fusion of matching scores of each iris.
During the training phase, the authors applied different \gls*{cnn} configurations and architectures, and chose the best one based on validation set results.
A $100$\% rank-1 recognition rate was obtained in SDUMLA-HMT, CASIA-Iris-V3, and IITD databases.
However, it is important to note that this methodology only works in a closed-world problem, since the matching score is based on the probability that an image belongs to a sample of a class known in the training phase.

Also using \gls*{nir} databases obtained in controlled environments, Nguyen et al.~\cite{Nguyen2018} demonstrated that generic descriptors using deep learning are able to represent iris features.
The authors compared five \gls*{cnn} architectures trained in the ImageNet database~\cite{Imagenet2009}.
The \glspl*{cnn} were used, without fine-tuning, for the feature extraction of normalized iris images.
Afterward, a simple multi-class \gls*{svm} was applied to perform the classification (identification). 
Promising results were presented in LG2200 (ND-CrossSensor-Iris-2013) 
and CASIA-Iris-Thousand databases, where all the architectures report better accuracy recognition than the baseline feature descriptor~\cite{Daugman2004}.

Other applications with iris images include spoofing detection~\cite{Menotti2015}, recognition of mislabeled left and right iris images~\cite{Du2016}, liveness detection~\cite{He2016b}, iris location~\cite{severo2018benchmark}, gender classification~\cite{Tapia2017} and sensor model identification~\cite{Marra2017}.

Also one can find the application of deep learning to the periocular and sclera regions, using images captured in uncontrolled environments~\cite{Luz2017, Proenca2018, lucio2018fully}. 

In all works found in the literature that apply deep learning for iris recognition, the input image used is the normalized one, where a previous iris location and segmentation is also required before the normalization process.
In this work, we evaluate the use of different input images for learning deep representations for iris recognition (verification).
The input images were created using three preprocesses: iris delineation, iris segmentation for noise removal and iris normalization.

\section{Protocol and Database}
\label{sec:protocol}

In this section, we describe the experimental protocol and database proposed in the NICE.II competition, which was used to evaluate the proposed methodology in this paper and in some of the related works.


The first iris recognition competition created specifically with images obtained in visible spectrum under uncontrolled environments is the Noisy Iris Challenge Evaluation (NICE).
The NICE competition is separated into two phases.
The first one, called NICE.I~(2008), was carried out with the objective of evaluating techniques for noise detection and segmentation of iris images.
In the second competition, NICE.II~(2010), the encoding and matching strategies were evaluated.
The NICE images were selected as a subset of images of a larger database, the UBIRIS.v2 \cite{Proenca2010}, which in turn comprises of $11$,$102$ images and $261$ individuals.


The main goal of NICE.I~\cite{Proenca2012} was to answer the following question: \enquote{Is it possible to automatically segment a small target as the iris in unconstrained data (obtained in a non-cooperative environment)?}.
The competition was attended by $97$ research laboratories from $22$ countries, which received a database of $500$ images to be used as training data in the construction of their methodologies.
These $500$ iris images were made available by the organizer committee, along with segmentation masks.
The masks were used as ground truth to assess the performance of iris segmentation methodologies.
For the evaluation of the submitted algorithms, a new database containing $500$ iris images was used to measure the pixel-to-pixel agreement between the segmentation masks created by each participant and the masks manually created by the committee.
The performance of each submitted methodology was evaluated with the error rate, which gives the average proportion of correctly classified pixels.

In order to guarantee impartiality and evaluate only the results of the feature extraction and matching, all the participants of the second phase of NICE~(NICE.II) used the same segmented iris images, which were obtained with the technique proposed by Tan et al.~\cite{Tan2010}, winner of NICE.I~\cite{Proenca2012}.
The objective of NICE.II was to evaluate how different sources of image noise obtained in an uncontrolled environment may interfere in iris recognition.
The training database consisted of $1$,$000$ images along with the corresponding segmented iris masks. 
The task was to build an executable that received as input a pair of iris images and their respective masks, generating a file with the corresponding scores with dissimilarity of irises~($d$) as an output. 
The~$d$ metric follows some conditions:
\begin{enumerate}
\item $d(I, I) = 0$
\item $d(I_{1}, I_{2}) = 0 \Rightarrow I_{1} = I_{2}$
\item $d(I_{1}, I_{2}) + d(I_{2}, I_{3}) \geq d(I_{1}, I_{3})$
\end{enumerate}

\begin{figure}[H]
\centering
\begin{tabular}{ccccc}
	{\includegraphics[width=0.18\columnwidth]{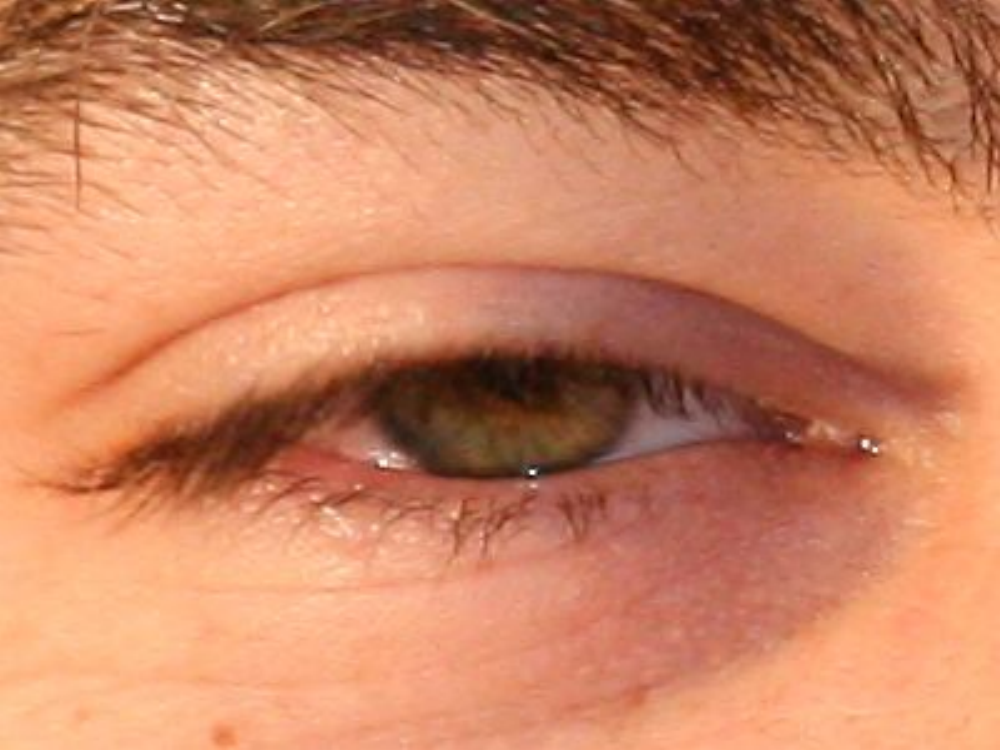}}
    {\includegraphics[width=0.18\columnwidth]{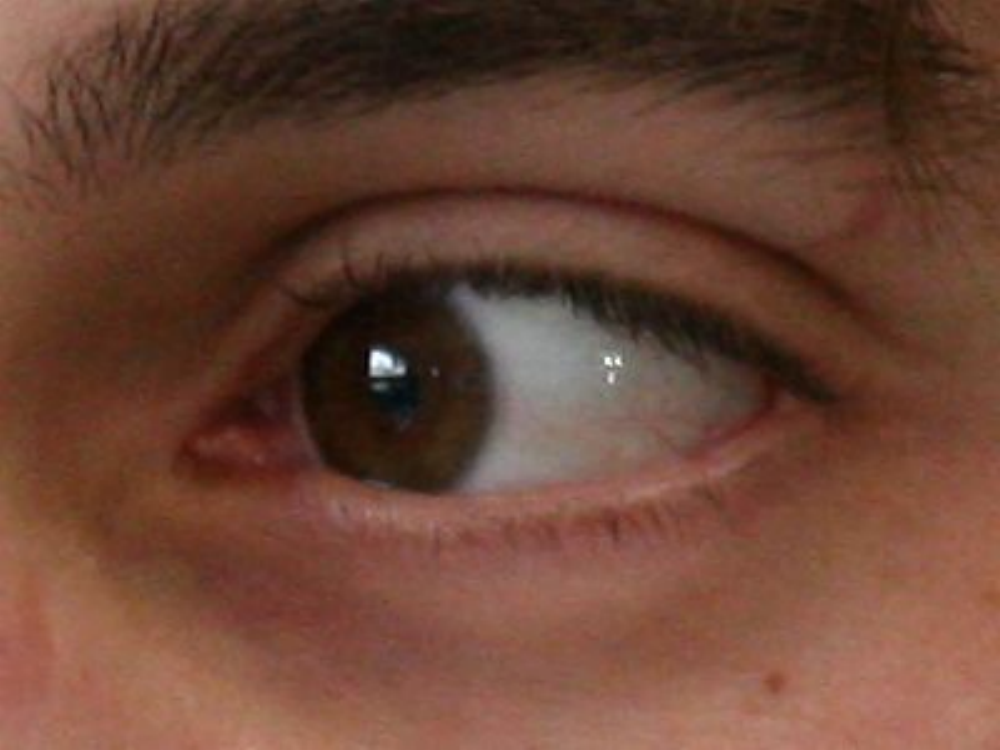}} 
    {\includegraphics[width=0.18\columnwidth]{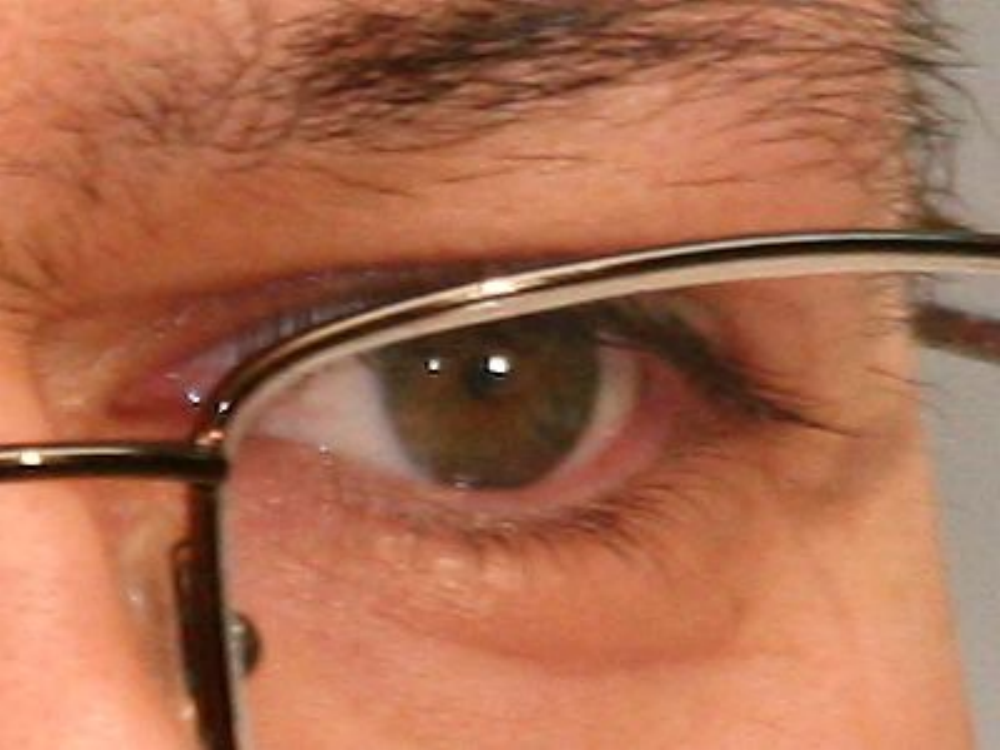}} 
    {\includegraphics[width=0.18\columnwidth]{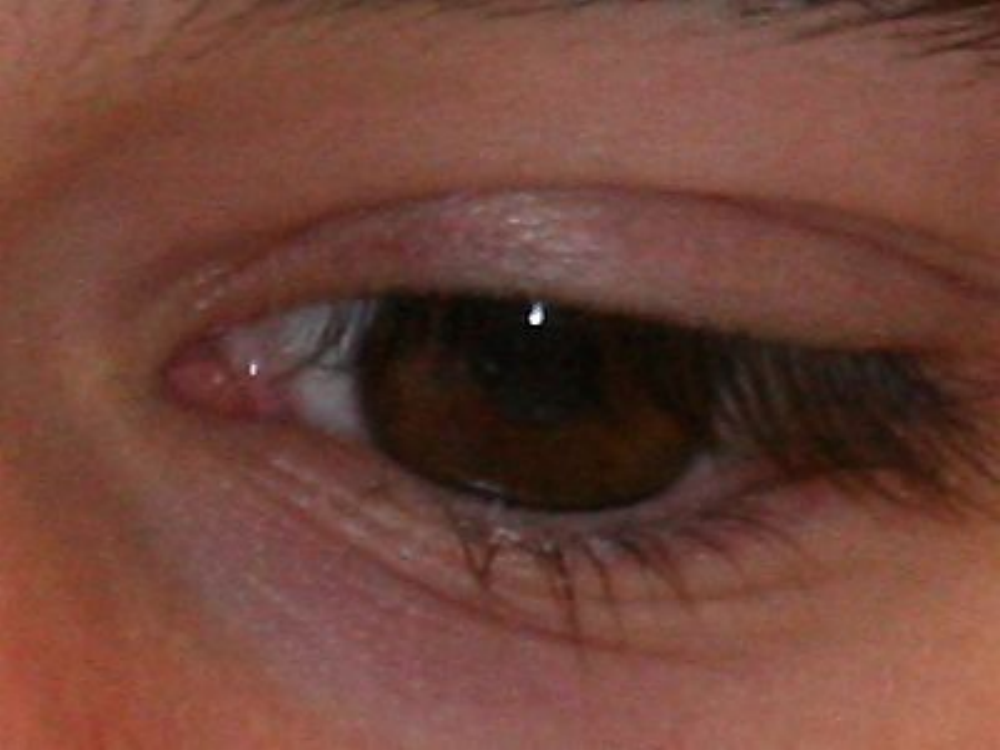}}
    {\includegraphics[width=0.18\columnwidth]{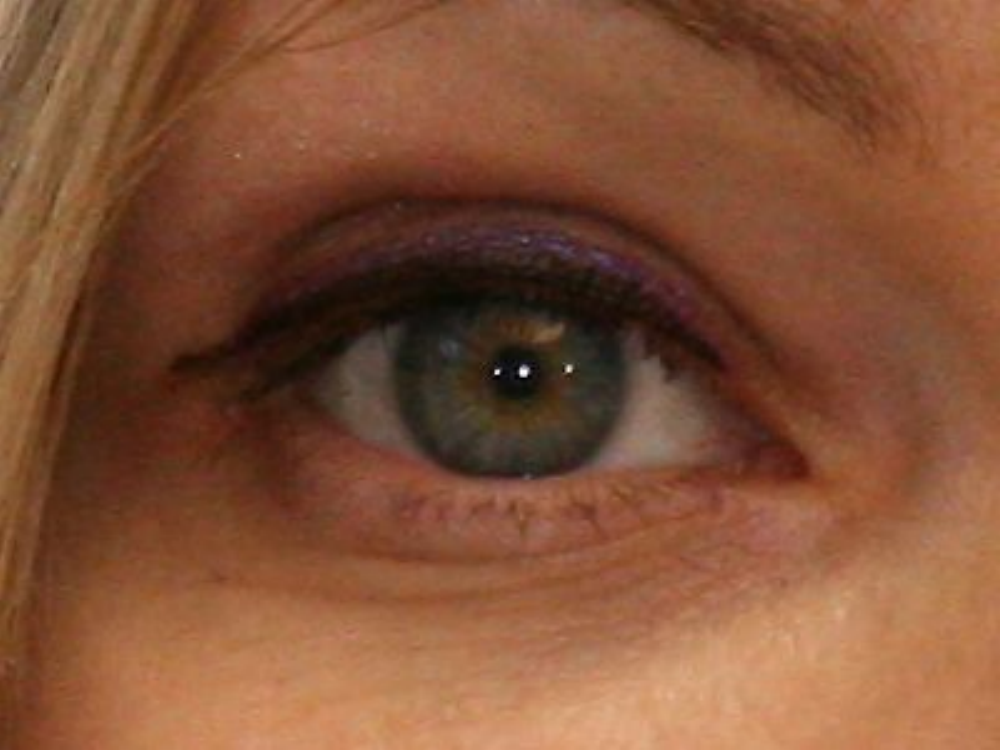}} \\ 

    {\includegraphics[width=0.18\columnwidth]{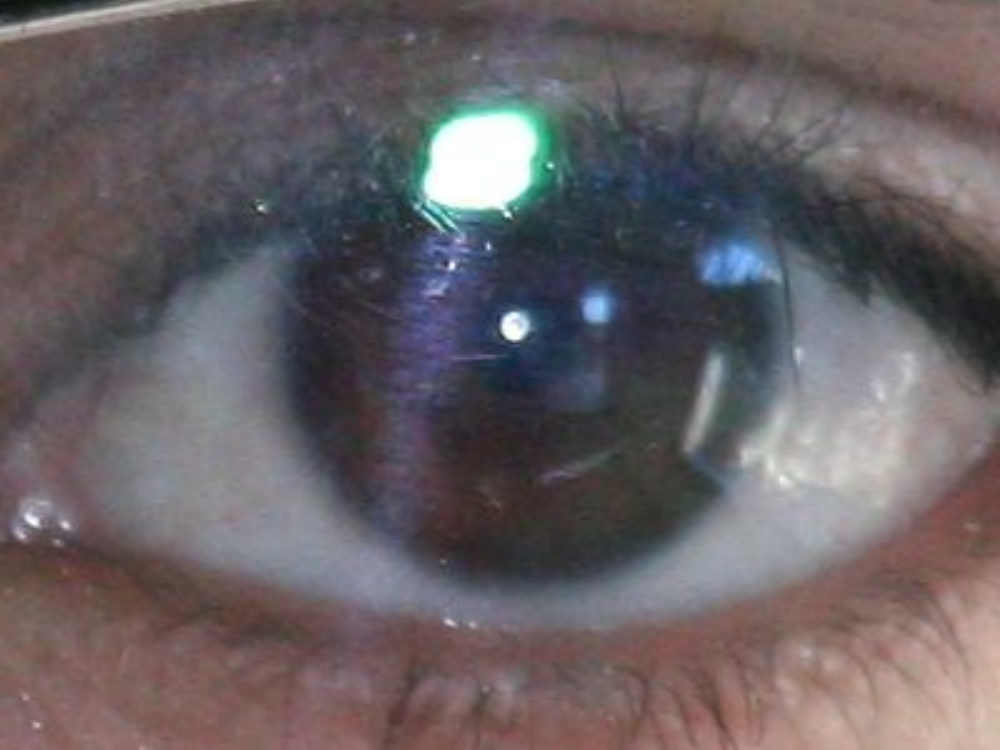}} 
    {\includegraphics[width=0.18\columnwidth]{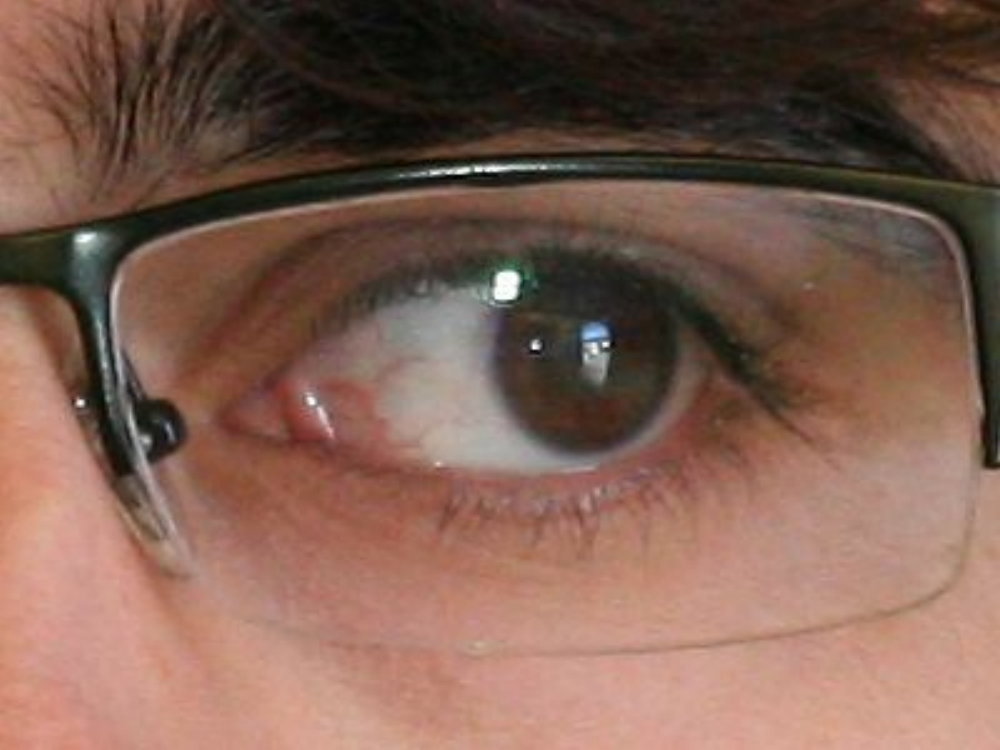}} 
    {\includegraphics[width=0.18\columnwidth]{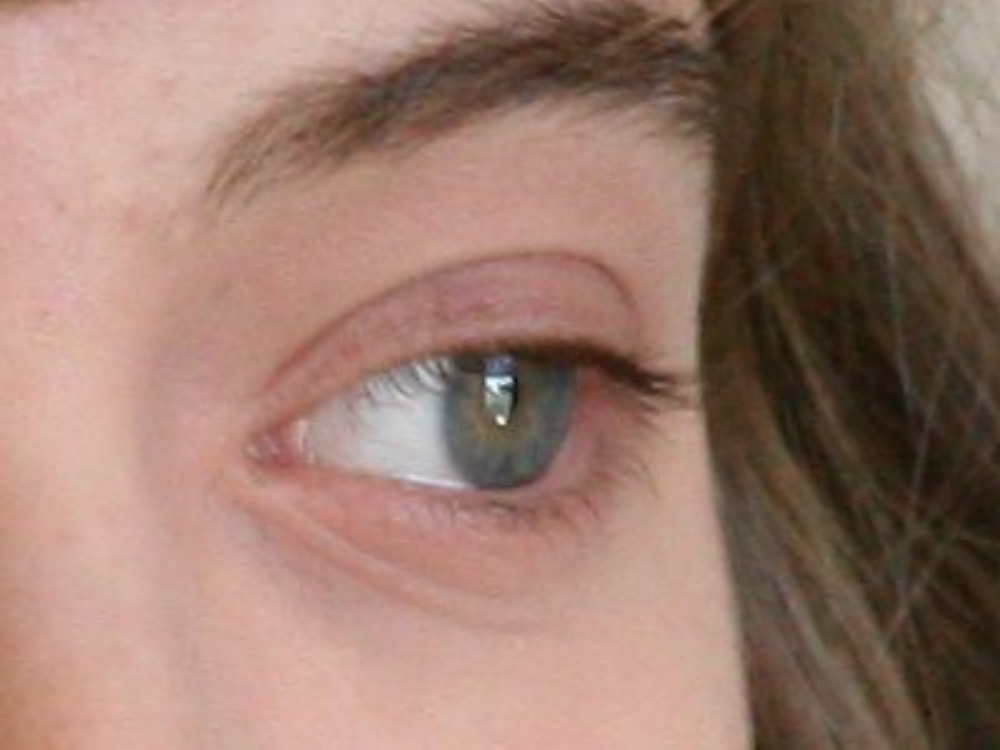}} 
    {\includegraphics[width=0.18\columnwidth]{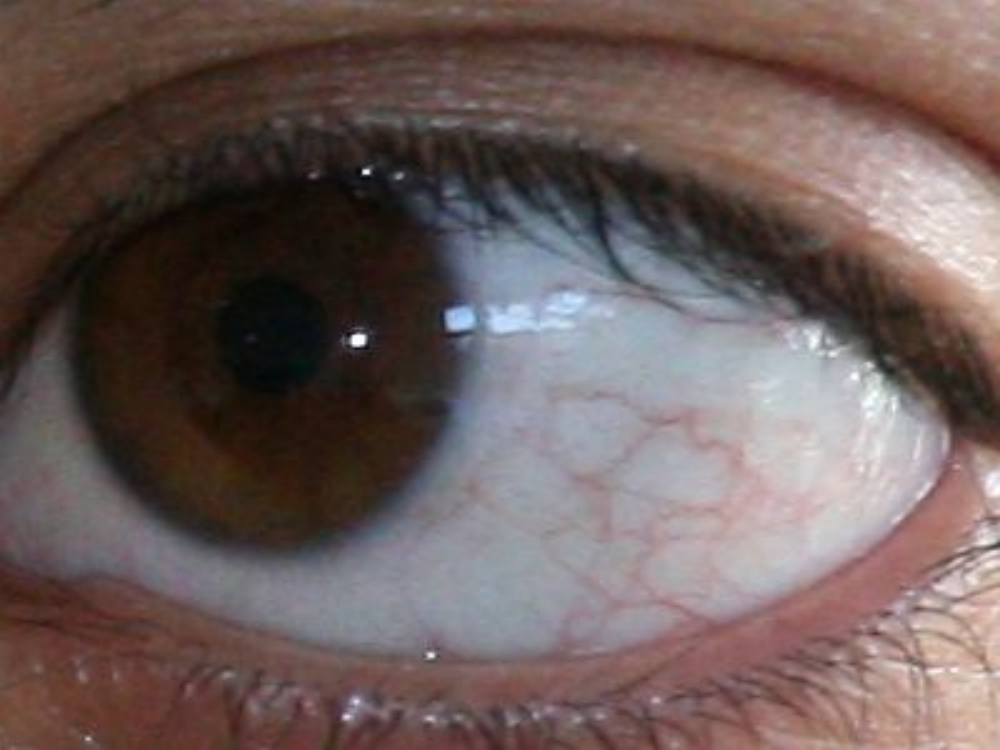}} 
    {\includegraphics[width=0.18\columnwidth]{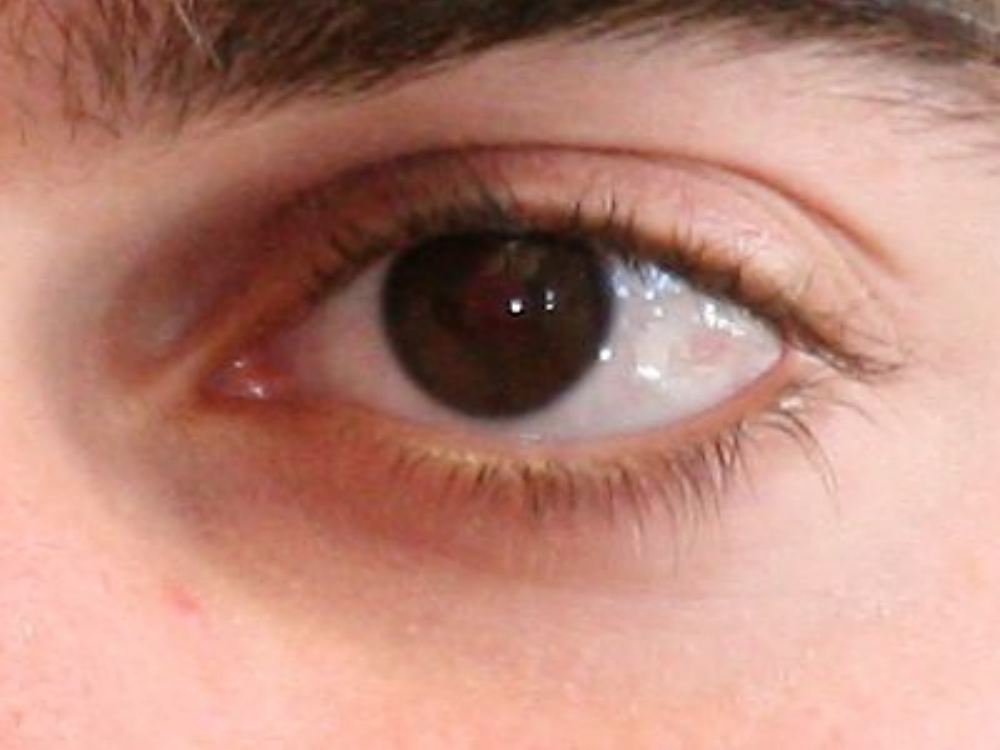}} \\

    {\includegraphics[width=0.18\columnwidth]{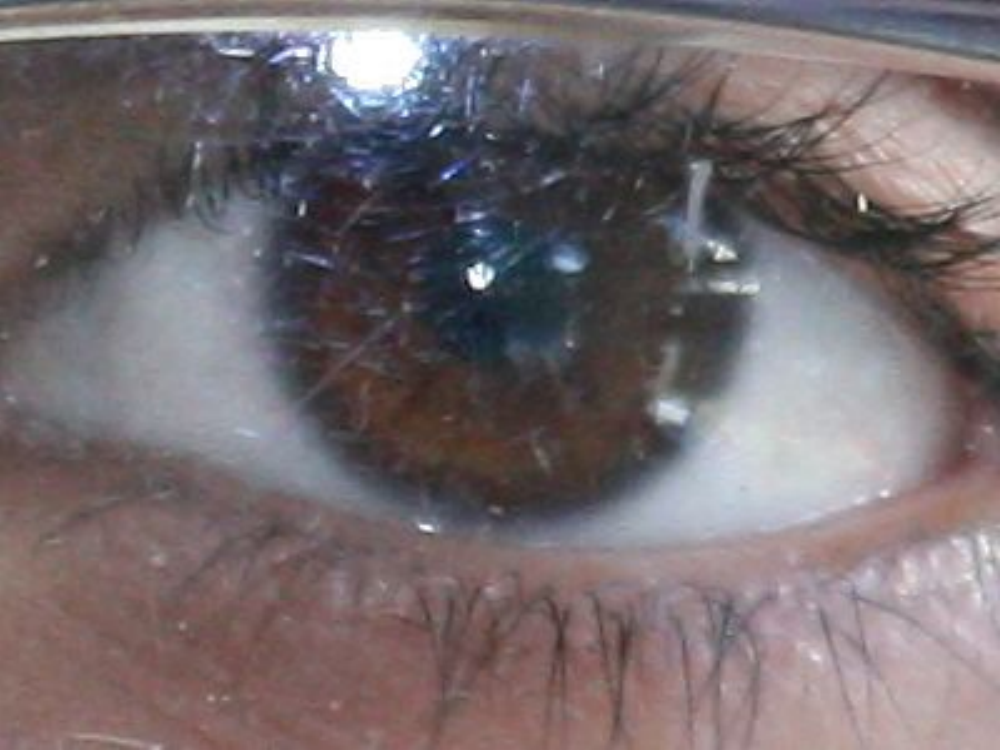}} 
    {\includegraphics[width=0.18\columnwidth]{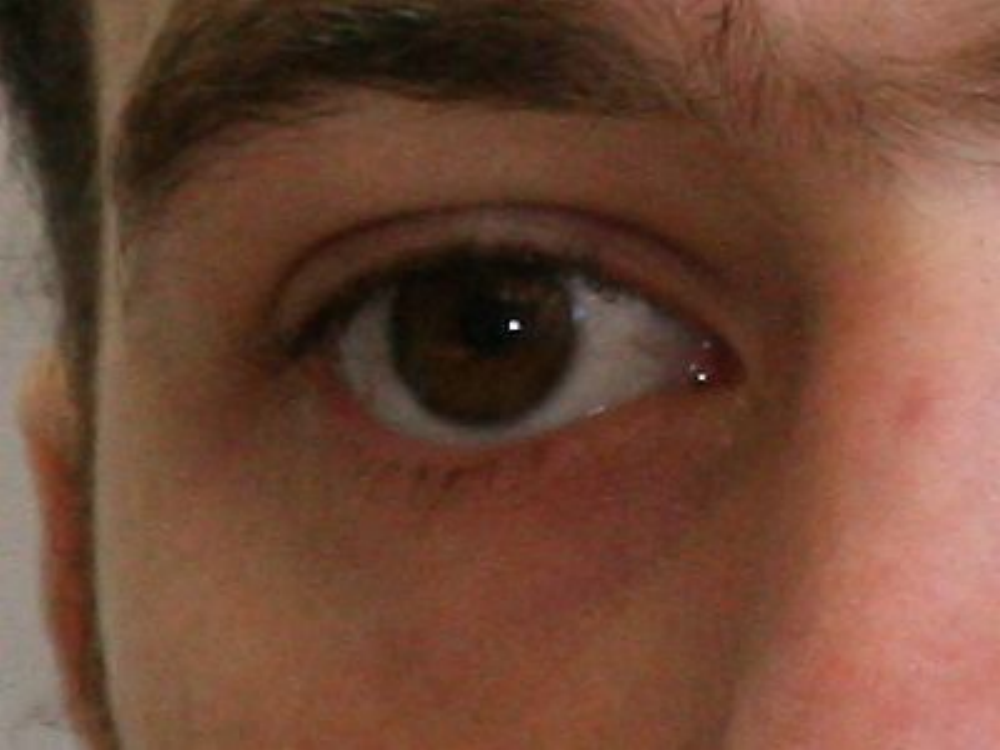}} 
    {\includegraphics[width=0.18\columnwidth]{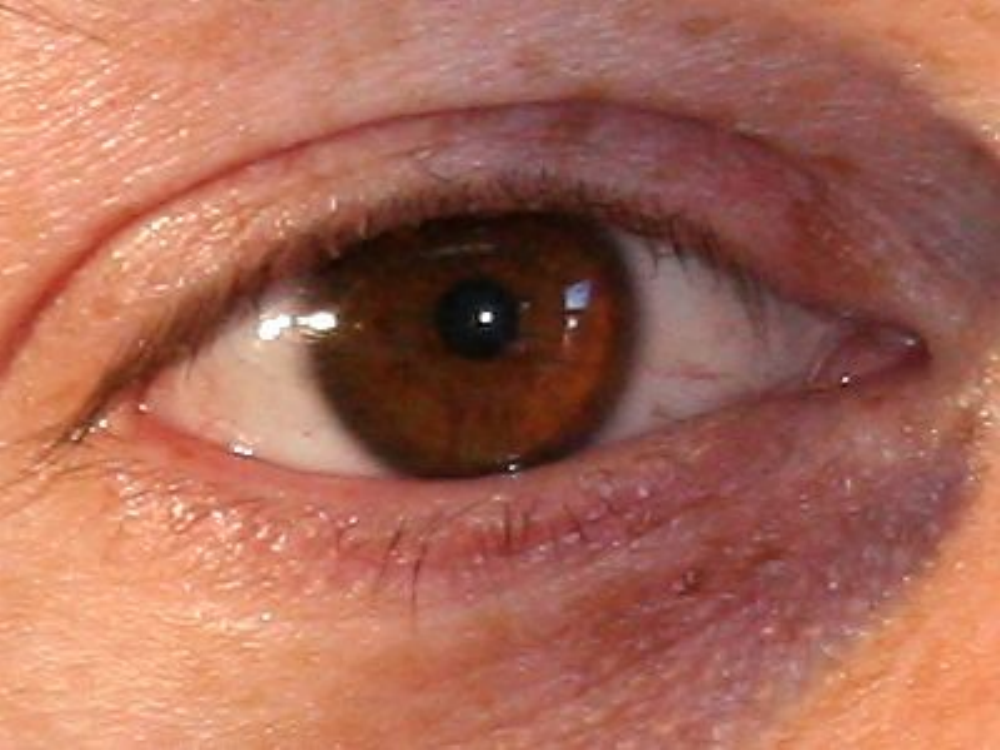}} 
    {\includegraphics[width=0.18\columnwidth]{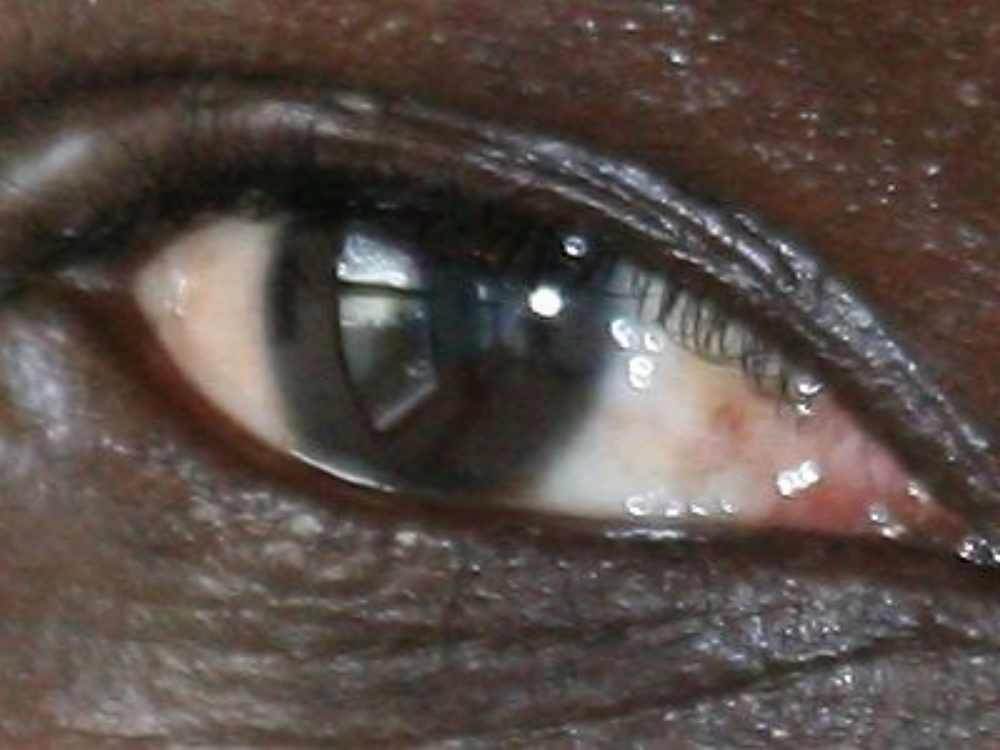}} 
    {\includegraphics[width=0.18\columnwidth]{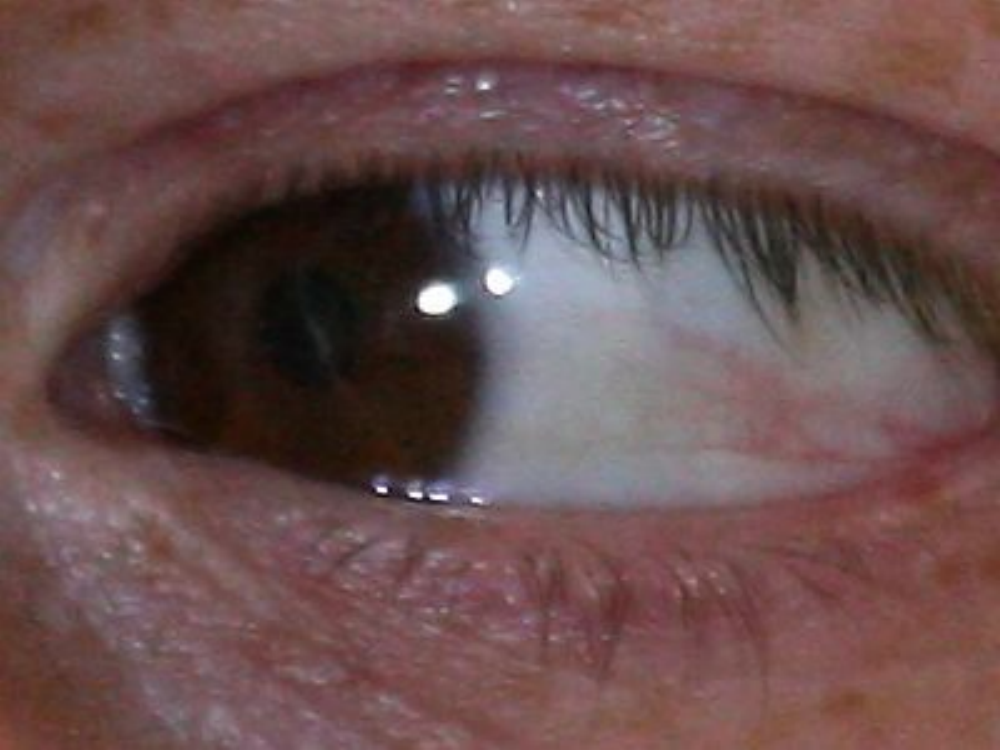}} \\ 

\end{tabular}
\caption{Samples from the NICE.II database.}
\label{fig:nicebd}
\end{figure}

For the evaluation of the methodologies proposed by the participants, another unknown database containing $1$,$000$ images and masks was employed.
Some samples randomly selected from these images are shown in Fig.~\ref{fig:nicebd}.
Consider $IM = \{I_{1},...,I_{n}\}$ as a set of iris images, $MA= \{M_{1},...,M_{n}\}$ as their targeting binary masks and $id(.)$ as the identity function on an image.
A comparison of one-against-all will return a match set $D^{I} = \{d^{i}_{1},...,d^{i}_{m}\}$ and non match $D^{E} = \{ d^{e}_{1},...,d^{e}_{k}\}$ of dissimilarity scores, respectively, for the cases where $id(I_{i}) = id(I_{j})$ and $id(I_{i}) \neq id(I_{j})$.
The evaluation of the algorithms was performed using the decidability scores $d'$~\cite{Daugman2003}.

The metric or index~$d'$ measures how well separated are two types of distributions, so the recognition error corresponds to the overlap area

\begin{equation}
d' = \frac{|\mu_{E} - \mu_{I}|}{\sqrt{\frac{1}{2} (\sigma^{2}_{I} + \sigma^{2}_{E})}}
\end{equation}

\noindent 
where the means of the two distributions are given by $\mu_{I}$ and $\mu_{E}$, and $\sigma_{I}$ and $\sigma_{E}$ represent the standard deviations.

Since the goal of the competition is the iris recognition in noisy images obtained in an uncontrolled environment, the methodology presented in this paper focuses only on the features extraction and matching using iris information.
As shown in previous studies~\cite{Tan2012,Santos2012,Ahmed2016,Ahmed2017}, promising results were achieved using the fusion of iris and periocular modalities, and since this fusion depends on the quality of each modality, it is interesting to study each one of them independently, i.e., using only iris or periocular information.


Considering that the decidability metric measures how discriminating the extracted features are, we have used it to compare our method with the state of the art.
We also report the \gls*{eer}, which is based on the \gls*{far} and the \gls*{frr}.
The \gls*{eer} is determined by the intersection point of \gls*{far} and \gls*{frr} curves.
All reported results were obtained in the official test set of the NICE.II competition.

\section{Proposed Approach}
\label{sec:approach}

\begin{figure*}[!htb]
\begin{center}
 \includegraphics[width=.95\linewidth]{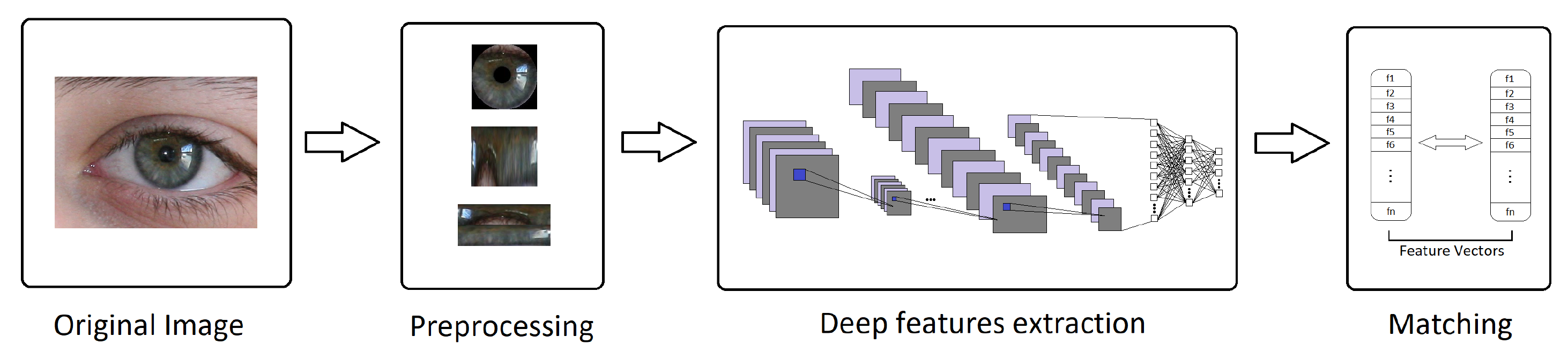}
\end{center}  
\caption{The proposed approach.}
\label{fig:proposed}
\end{figure*}

The proposed approach consists of three main stages: image preprocessing, feature extraction and matching, as shown in Fig.~\ref{fig:proposed}.
In the first stage, segmentation and normalization techniques are applied.
We perform data augmentation in the preprocessed images to increase the number of training samples.
The feature extraction is performed with two \gls*{cnn} models, which were fine-tuned using the original images, and the images generated through data augmentation.
Finally, the matching is performed using the cosine distance.
These processes are best described throughout this
section.

\subsection{Image preprocessing}

\begin{figure}[!htb]
\centering
\begin{tabular}{ccccc}

    \multirow{4}{*}{\includegraphics[width=0.40\columnwidth]{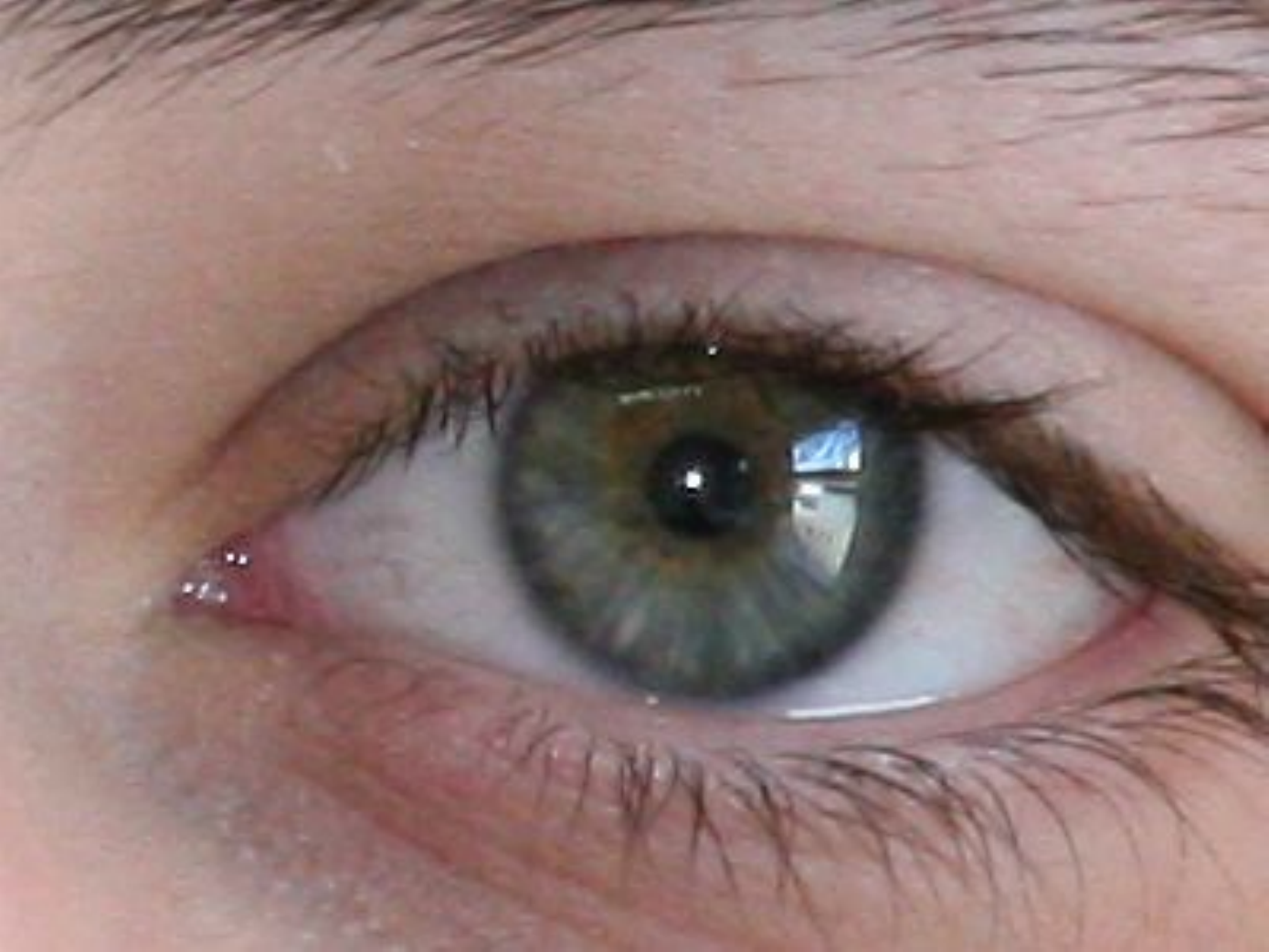}} 
    &{\includegraphics[width=0.20\columnwidth]{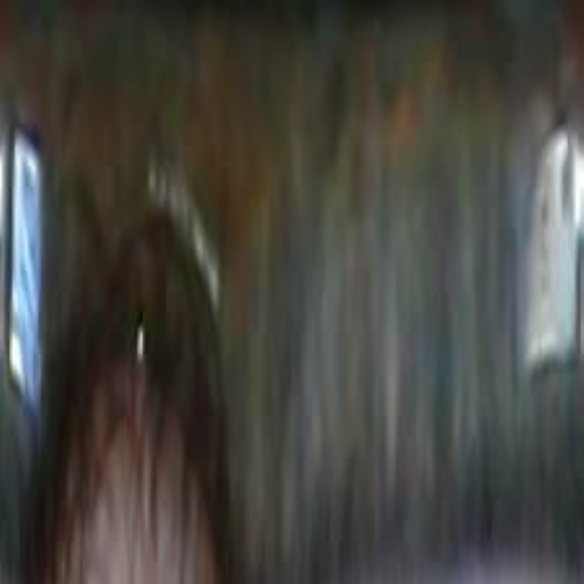}} 
    {\includegraphics[width=0.20\columnwidth]{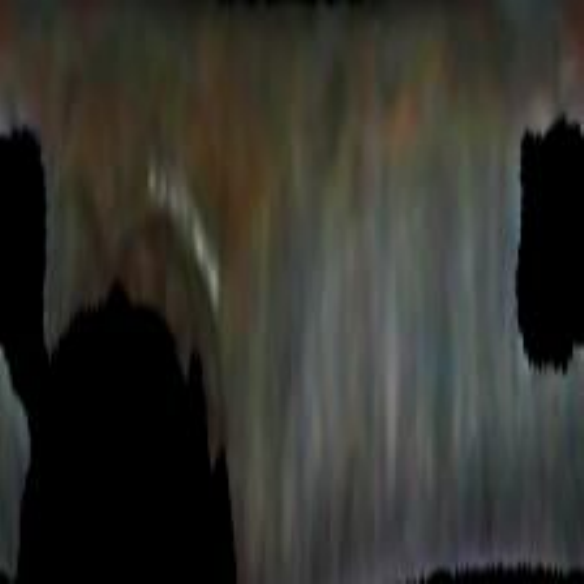}}  
\\
    &{\includegraphics[width=0.20\columnwidth]{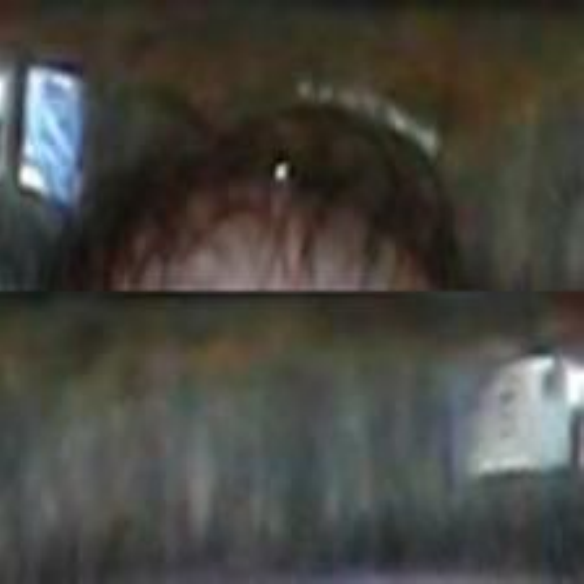}}
    {\includegraphics[width=0.20\columnwidth]{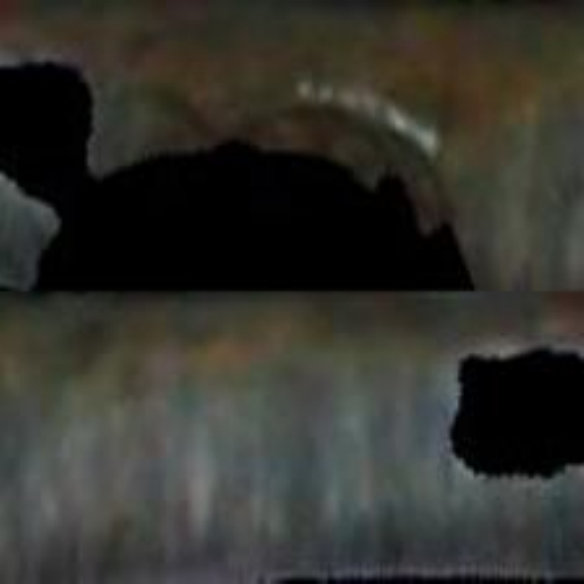}}  
\\
    &{\includegraphics[width=0.20\columnwidth]{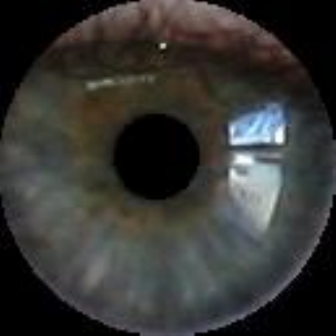}}
    {\includegraphics[width=0.20\columnwidth]{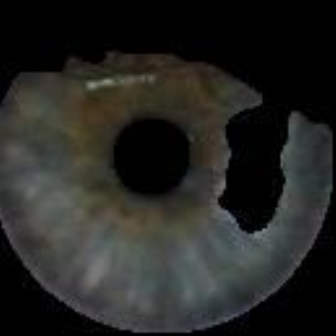}}
\\
    
    \footnotesize
    \parbox{.40\linewidth}{\centering (original image)} &
	\footnotesize
    \parbox{.20\linewidth}{\centering (a)} 
	\footnotesize
	\parbox{.20\linewidth}{ \centering (b)}

\end{tabular}
\caption{Input Image. (a) and (b) show, respectively, non-segmented and segmented images for noise removal from the Nice.II database. From top to bottom, images with the aspect ratios of $8:1$, $4:2$ are shown, as well as non-normalized images.}
\label{fig:inputrep}
\end{figure}

In order to analyze the impact of the preprocessing, six different input images were generated from the original iris image, as shown in Fig.~\ref{fig:inputrep}.
In the first image scheme, irises are normalized with the standard rubber sheet model~\cite{Daugman1993} using an $8:1$ aspect ratio ($512 \times 64$ pixels).
In the second image scheme, the iris images are also normalized, however, instead of the standard $8:1$ ratio they were rearranged in a $4:2$ ratio ($256 \times 128$ pixels), so that less interpolation is employed in the resizing process.
In the third and last schemes, no normalization is performed, applying only the delineated iris images as input to the models.
Non-normalized images have different sizes, according to the image capture distance.
The impact of the segmentation technique for noise removal was also evaluated in all representations.
Note that all the iris images used as inputs for the feature representation models are resized using bi-cubic interpolation to $224\times224$ pixels.

\newcolumntype{C}[1]{>{\centering\let\newline\\\arraybackslash\hspace{0pt}}m{#1}}

The normalization through the rubber sheet model~\cite{Daugman1993} aims to obtain invariance with respect to size and pupil dilatation.
In the NICE.II database, the main problem is the difference of the iris size due to distances in the image capture.
It is important to note that in non-normalized images, we use an arc delimitation preprocessing (i.e., two circles, an outer and an inner), based on the iris mask.

The segmentation process to noise removal was performed using the masks provided along with the database.
These masks were obtained with the methodology proposed by Tan et al.~\cite{Tan2010}, winner of NICE.I~\cite{Proenca2012}.

Considering these problems, the proposed approach aims to analyze the impact of non-normalization and non-segmentation of iris images to extract deep features.

\subsection{Data Augmentation}

Since the training subset has only $1$,$000$ images belonging to $171$ classes, it is important to apply data augmentation techniques to increase the number of training samples.
The fine-tuning process can result in a better generalization of the models with more images.
In this sense, we rotate the original images at specific angles.

\begin{figure}[!htb]
\centering
\begin{tabular}{ccc}

    {\includegraphics[width=20mm]{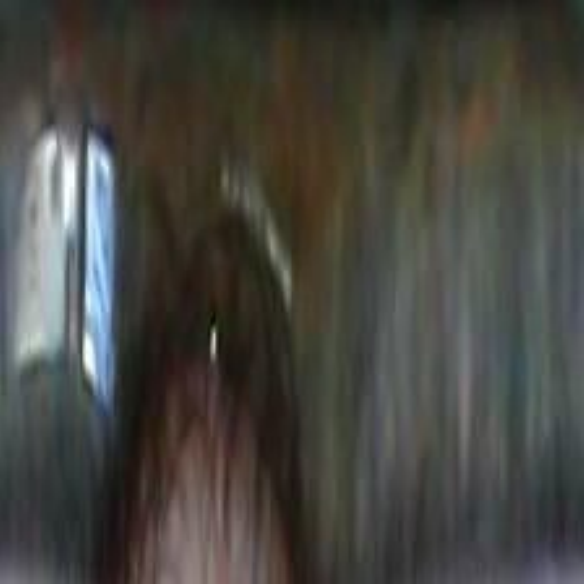}} 
    {\includegraphics[width=20mm]{figs/C10I486P1Norm224x224NOSeg.pdf}} 
    {\includegraphics[width=20mm]{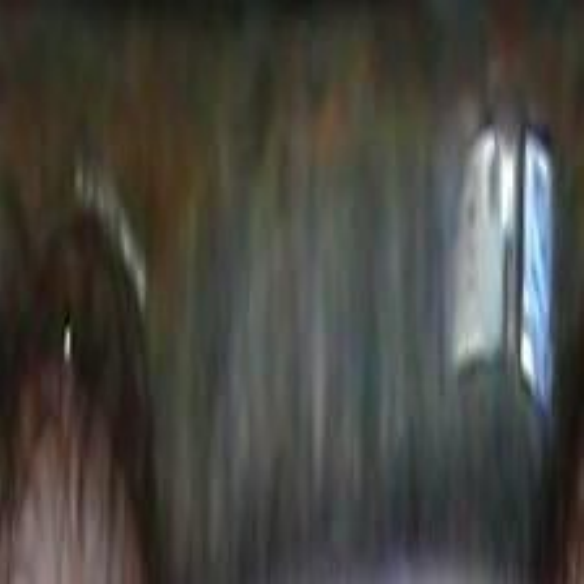}} \\

    {\includegraphics[width=20mm]{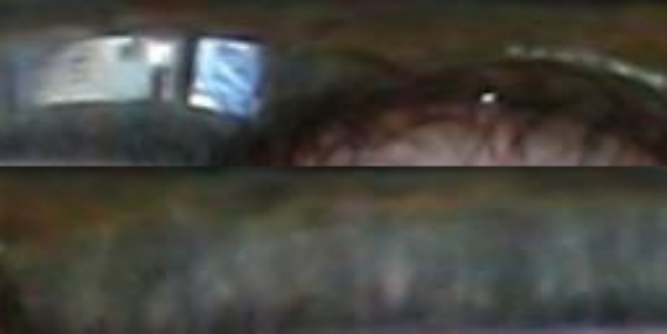}} 
    {\includegraphics[width=20mm]{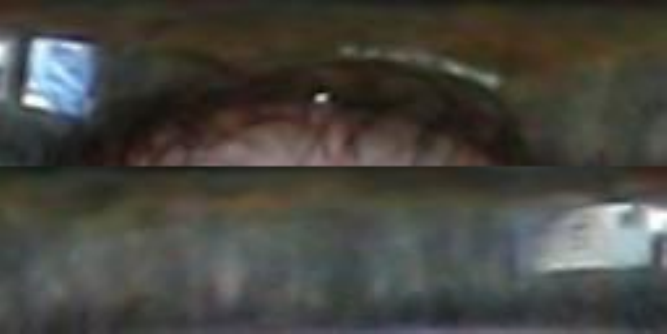}}
    {\includegraphics[width=20mm]{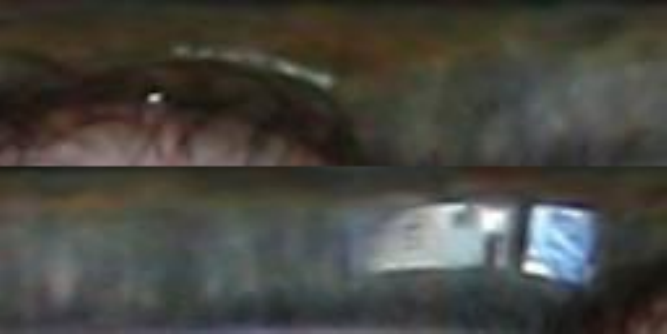}} \\
    
    {\includegraphics[width=20mm]{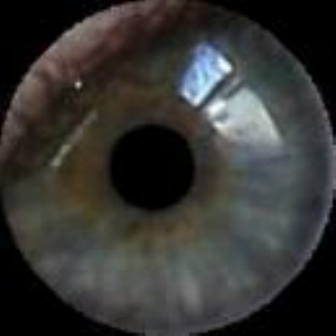}} 
    {\includegraphics[width=20mm]{figs/C10I486P1NONormNOSeg.pdf}}
    {\includegraphics[width=20mm]{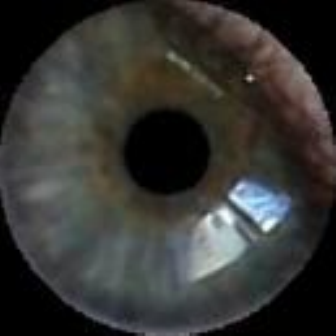}} \\
    
    \footnotesize
    \parbox{.23\linewidth}{\centering (a)} 
	\footnotesize
	\parbox{.23\linewidth}{ \centering (b)}
    \footnotesize
	\parbox{.23\linewidth}{ \centering (c)} \\

\end{tabular}
\caption{Data augmentation samples: (a)~\ang{-45} rotated images, (b)~original images and (c)~\ang{45} rotated images.}
\label{fig:dataaug}
\end{figure}

The ranges of angles used were: \ang{-15} to \ang{+15}, \ang{-30} to \ang{+30}, \ang{-45} to \ang{+45}, \ang{-60} to \ang{+60}, \ang{-90} to \ang{+90} and \ang{-120} to~\ang{+120}.
For each range, the aperture is proportional, generating $4$, $6$ and $8$ images for each original image, respectively.
For example, considering the range \ang{-60} to \ang{+60} with $6$ apertures, for each original image, another six were generated rotating \ang{-60}, \ang{-40}, \ang{-20}, \ang{+20}, \ang{+40} and~\ang{+60}.
Performing the validation of all these data augmentation methods on all input images, we determined (based on accuracy and loss) that the best range was \ang{-60} to \ang{+60} with $6$ apertures.
These parameters were applied to perform the data augmentation in the training set, totaling $7$,$000$ images. Some samples generated by data augmentation can be seen in Fig.~\ref{fig:dataaug}.

\subsection{Convolutional Neural Network Model}

For feature extraction of the iris images, the fine-tuning of two \gls*{cnn} models trained for face recognition were applied.
The first model, called VGG, proposed in~\cite{Omkar2015} and used in~\cite{Luz2017} for periocular recognition, has an architecture composed of convolution, activation~(ReLu), pooling and fully connected layers.
The second model (i.e., Resnet-50), proposed in~\cite{He2016a} and trained for face recognition~\cite{Cao2017}, has the same operations as VGG with the difference of being deeper and considering residual information between the layers.

For both models, we use the same architecture modifications and parameters described in~\cite{Luz2017}.
In the training phase, the last layer (used to predict) was removed and two new layers were added.
The new last layer, used for classification, is composed by $171$ neurons, where each one corresponds to a class in the NICE.II training set and has a softmax-loss function.
The layer before that is a fully-connected layer with $256$ neurons used to reduce feature dimensionality.

The training set was divided into two new subsets with $80$\% of the data for training and $20$\% for validation.
Two learning rates were used for $30$ epochs, $0.001$ for the first $10$ epochs and $0.0005$ for the remaining~$20$. Other parameters include momentum~=~$0.9$ and batch size~=~$48$.
The number of epochs used for training was chosen based on the experiments carried out in the validation set (highest accuracy and lowest loss).
For training (fine-tuning) the \gls{cnn} models, the Stochastic Gradient Descent (SGD) optimizer was used.
Similarly to~\cite{Long2015, Luz2017}, we do not freeze the weights of the pre-trained layers during training to perform the fine-tuning.
The last layer of each model was removed and the features were extracted on the new last layer.

\subsection{Matching}

The matching was performed using the verification protocol proposed in the NICE.II competition.
For this, the all against all approach was applied in the NICE.II test set, generating $4$,$634$ intra-class pairs and $494$,$866$ inter-class pairs.


The cosine metric, which measures the cosine of the angle between two vectors, was applied to compute the difference between feature vectors.
This metric is used in information retrieval~\cite{van2012metric} due to its invariance to scalar transformation.
The cosine distance metric is represented by
\begin{equation}
d_{c}(A,B) = 1 - \frac{\sum_{j=1}^{N}A_{j}B_{j}}{\sqrt{\sum_{j=1}^{N}A_{j}^2} \sqrt{\sum_{j=1}^{N}B_{j}^2}}
\end{equation}
where $A$ and $B$ stand for the feature vectors.
We also employed other distances such as Euclidean, Mahalanobis, Jaccard and Manhattan.
However, due to its best performance, only the cosine distance is reported.

\section{Results}
\label{sec:results}

In this section, we present the results of the experiments for validating our proposal using the test set from the Nice.II competition database.
Initially, the impact of the proposed data augmentation techniques is shown using non-segmented iris images.
Then, we analyze the impact of both iris segmentation and iris delineation.
Finally, we compare the best results obtained by our approaches with the state of the art.
In all subsections, the impact of  normalization is also analyzed.
Note that in all experiments, the mean and standard deviation values from $30$ runs are reported.
For analyzing the different results, we perform statistical paired t-tests at a significance level $\alpha = 0.05$.

\subsection{Data Augmentation}

In the first analyses, we evaluate the impact of the data augmentation.
For ease of analysis, all iris image employed in this initial experiment may contain noise in the iris region, i.e. no segmentation preprocessing is applied.
As shown in Table~\ref{tab:expdataaug}, in all cases where data augmentation is used, the decidability and \gls*{eer} values improved with statistical difference.
Note that the models trained with data augmentation reported smaller standard deviation.
In general, it is also observed that non-normalization yielded better results than $8:1$ and $4:2$ normalization schemes for both trained models, i.e. VGG and ResNet-50.

It is worth noting that the largest differences occurred in the non-normalized inputs, with greater impact specifically in the ResNet-50 model, where the mean \gls*{eer} dropped $7.53\%$ and the decidability improved $0.6361$ when applying data augmentation.

\subsection{Segmentation}

In the second analysis, the impact of the segmentation for noise removal is evaluated.
For such aim, two models are trained (fine-tuned): using segmented and non-segmented images, all with data augmentation.

As can be seen in Table~\ref{tab:expdataaug}, for the VGG model segmentation has improved the results.
On the other hand, for the ResNet-50 model, the non-segmented images have presented better results.
For both models, statistical difference is achieved in two situations and in another one (light cyan color) there is no statistical difference.

Regarding the better results achieved by the ResNet-50 models when using non-segmented images, we hypothesized that this might be related to the fact that the ResNet-50 architecture uses residual information and it is deeper compared to VGG.
Thus, some layers of ResNet-50 might be responsible for extracting discriminant patterns present in regions that were occluded in the segmented images, but not in non-segmented ones.
Moreover, in segmented images, black regions (zero values) were employed for representing noise regions, and no special treatment was given for those regions.

It is noteworthy that segmentation is a complex process and might impact positively or negatively.
However, as the best results here were achieved by the ResNet-50 models when using non-segmented images, we state that using the suitable representation model the segmentation preprocessing can be disregarded.


Once again, non-normalization showed to provide better results in all scenarios, being  more expressive here than in the data augmentation analysis.

\begin{table}[!ht]
\centering
\caption{Impact of the Data Augmentation (DA) and Segmentation (Seg.) on the effectiveness of iris verification for VGG and ResNet-50 networks.}
\label{tab:expdataaug}
\resizebox{\columnwidth}{!}{
\begin{tabular}{@{}lc@{ }c@{ }ccc@{}}
\toprule
\centering{Network} & Norm.              & {DA}                 & {Seg.}             & {EER (\%)}                & {Decidability} \\
\midrule
VGG                 & $8:1$              &                      &                    & $26.19\pm1.95$            & $1.3140\pm0.1246$    \\
VGG                 & $8:1$              & \checkmark           &                    & $23.63\pm1.33$            & $1.4712\pm0.0881$ \\
VGG                 & $8:1$              & \checkmark           & \checkmark         & $22.58\pm1.07$            & $1.5437\pm0.0697$    \\
ResNet-50           & $8:1$              &                      &                    & $24.38\pm1.41$            & $1.4297\pm0.0916$    \\
ResNet-50           & $8:1$              & \checkmark           &                    & $19.18\pm0.75$            & $1.7988\pm0.0552$    \\
ResNet-50           & $8:1$              & \checkmark           & \checkmark         & $20.68\pm1.39$            & $1.6801\pm0.1071$    \\
\midrule
VGG                 & $4:2$              &                      &                    & $24.77\pm1.42$            & $1.4127\pm0.1001$    \\
VGG                 & $4:2$              & \checkmark           &                    & $18.74\pm0.89$            & $1.8527\pm0.0712$    \\
VGG                 & $4:2$              & \checkmark           & \checkmark         & $18.00\pm0.93$            & $1.9055\pm0.0750$    \\
ResNet-50           & $4:2$              &                      &                    & $22.78\pm1.22$            & $1.5307\pm0.0853$    \\
\rowcolor{lightblue}
ResNet-50           & $4:2$              & \checkmark           &                    & $17.11\pm0.53$            & $1.9822\pm0.0482$    \\
\rowcolor{lightblue}
ResNet-50           & $4:2$              & \checkmark           & \checkmark         & $17.44\pm0.85$            & $1.9450\pm0.0803$    \\
\midrule
VGG                 & Non-Norm           &                      &                    & $23.32\pm1.10$            & $1.4891\pm0.0740$    \\
\rowcolor{lightblue}
VGG                 & Non-Norm           & \checkmark           &                    & $17.49\pm0.90$            & $1.9529\pm0.0760$   \\
\rowcolor{lightblue}
\textbf{VGG}        & \textbf{Non-Norm}  & \textbf{\checkmark}  & \textbf{\checkmark}& \boldmath{$17.48\pm0.68$} & \boldmath{$1.9439\pm0.0589$}     \\
ResNet-50           & Non-Norm           &                      &                    & $21.51\pm0.97$            & $1.6119\pm0.0677$    \\
\textbf{ResNet-50}  & \textbf{Non-Norm}  & \textbf{\checkmark}  &                    & \boldmath{$13.98\pm0.55$} & \boldmath{$2.2480\pm0.0528$}    \\
ResNet-50           & Non-Norm           & \checkmark           & \checkmark         & $14.89\pm0.78$            & $2.1781\pm0.0794$     \\

\bottomrule
\end{tabular}}
\end{table}

\subsection{Delineation}

\begin{figure}[!htb]
\centering
\begin{tabular}{cc}

	{\includegraphics[width=0.30\columnwidth]{figs/C10I486P1NONormNOSeg.pdf}}   
    {\includegraphics[width=0.30\columnwidth]{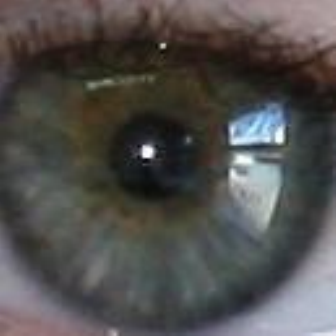}} \\
    
    \footnotesize
    \parbox{.30\linewidth}{\centering (a)} 
	\footnotesize
	\parbox{.30\linewidth}{ \centering (b)} \\

\end{tabular}
\caption{Input images: (a) delineated iris and (b) non-delineated iris / bounding box version.}
\label{fig:delin}
\end{figure}

Here we evaluated the impact on the recognition of using a usual delineated iris image and a non-delineated iris image, i.e., applying only the \textit{squared} iris bounding box as input to the deep feature extractor.
In both situations, non-normalized and non-segmented images are used.
A delineated iris image and its corresponding bounding box (or non-delineated) are shown in Fig.~\ref{fig:delin}. 

\begin{table}[!ht]
\centering
\caption{Comparison of delineated and non-delineated iris images. Both with no segmentation (for noise removal) and normalization.}
\label{tab:expbbox}
\begin{tabular}{@{}lccc@{}}
\toprule
\centering {Method}  & Delineated  & {EER (\%)}     & {Decidability} \\
\midrule
VGG                  &\checkmark   & $17.49\pm0.90$ & $1.9529\pm0.0760$    \\
VGG                  &             & $17.52\pm0.98$ & $1.9652\pm0.0790$    \\
Resnet-50            &\checkmark   & $13.98\pm0.55$ & $2.2480\pm0.0528$    \\
Resnet-50            &             & $14.26\pm0.47$ & $2.2304\pm0.0542$    \\

\bottomrule
\end{tabular}
\end{table}

The comparison of the results of this analysis is shown in Table~\ref{tab:expbbox}.
Although the results reported by delineated iris images are better, there is no statistical difference. 
From this result, we state that the iris bounding box can be used as input for deep representation without the iris delineating (a.k.a. location) preprocessing.

Considering that the bounding box is not pure iris, comparison with other iris recognition methods may not be fair, since there may be discriminant patterns that have been extracted from regions outside the iris.
Therefore, our methodology was compared with the state of the 
art using delineated iris images.


\subsection{The state of the art} 

At last, the results attained with our models using non-normalized, non-segmented, and delineated iris images are compared with the state-of-the-art approaches and it is shown in Table~\ref{tab:expsoa}.

These experiments showed that the representations learned using deep models perform better the iris verification task on the NICE.II competition  when  the preprocessing steps of normalization and segmentation (for noise removal) are removed, outperforming the state-of-the-art method, which uses preprocessed images.

\begin{table}[!ht]
\centering
\caption{Results on the NICE.II competition database. Comparison of the state of the art with the results achieved by our proposed approaches using non-normalized, non-segmented, and delineated iris images.}
\label{tab:expsoa}
\begin{tabular}{@{}lcc@{}}
\toprule
\centering {Method}         & {EER (\%)}   & {Decidability} \\
\midrule
Wang et al\cite{Wang2012}   & $19.00$         & $1.8213$    \\
Proposed VGG                & $17.48$	      & $1.9439$    \\
\textbf{Proposed ResNet-50} & \boldmath{$13.98$} & \boldmath{$2.2480$}    \\

\bottomrule
\end{tabular}
\end{table}



\section{Conclusion}
\label{sec:conclusion}

In this paper we evaluated the impact of iris image preprocessing  for iris recognition  on unconstrained environments using deep representations.
Different combinations of (non-)normalized and (non-)segmented images as input for the system were evaluated.

Using these iris images, a fine-tuning process of two \gls*{cnn} architectures pre-trained for face recognition was performed.
These models were applied to extract deep representations.
The matching, on a verification protocol, was performed with the cosine metric.
A significant improvement in the results of both models was achieved using the proposed data augmentation approach.
For both models, non-normalized iris achieved a better result.
In addition, we verified that the use of non-delineated iris images is slightly worst than the ones attained when using delineated images, but no significant difference was reached.
However, for a fair comparison with a state-of-the-art method, we used only delineated images because they represent the pure irises.

The experiments showed that the models learned on the ResNet-50 architecture using non-segmented and non-normalized images reported the best results achieving a new state of the art in the NICE.II official protocol -- one of the most challenging databases on uncontrolled environments.

As future work, we intend to evaluate this approach in larger databases and the performance of other network architectures for feature extraction and also transfer learning from other domains than face recognition.
Dealing with noisy iris regions and the analysis of its impact on iris recognition is the plan for future work as well.

\section*{Acknowledgment}

The authors thank the \textit{Conselho Nacional de Pesquisa e Desenvolvimento}~(CNPq)  (\#~428333/2016-8 and \#~313423/2017-2) for the financial support and also gratefully acknowledge the support of NVIDIA Corporation with the donation of the Titan Xp GPU used for this research.



\balance


\bibliographystyle{IEEEtran}
\bibliography{example}
%
%


\end{document}